\documentclass[10pt,twocolumn,letterpaper]{article}

\usepackage{cvpr}              

\usepackage{graphicx}
\usepackage{amsmath}
\usepackage{amssymb}
\usepackage{booktabs}
\usepackage{acronym}
\DeclareMathOperator*{\argmax}{argmax}

\usepackage[pagebackref,breaklinks,colorlinks]{hyperref}

\usepackage[capitalize]{cleveref}
\crefname{section}{Sec.}{Secs.}
\Crefname{section}{Section}{Sections}
\Crefname{table}{Table}{Tables}
\crefname{table}{Tab.}{Tabs.}

\def\cvprPaperID{4900} 
\def\confName{CVPR}
\def\confYear{2022}

\acrodef{vqa}[VQA]{Visual Question Answering}
\acrodef{kbvqa}[KB-VQA]{Knowledge-Based Visual Question Answering}
\acrodef{okvqa}[OK-VQA]{Outside-Knowledge Visual Question Answering}

\begin{document}

\title{A Thousand Words Are Worth More Than a Picture: \\ Natural Language-Centric Outside-Knowledge Visual Question Answering}

\author{
Feng Gao\textsuperscript{1},
Qing Ping\textsuperscript{2}, 
Govind Thattai\textsuperscript{2},
Aishwarya Reganti \textsuperscript{2},
Ying Nian Wu \textsuperscript{1},
Prem Natarajan\textsuperscript{2}
\\
\textsuperscript{1}University of California, Los Angeles,  \textsuperscript{2}Amazon\\
{\tt\small f.gao@ucla.edu,}
{\tt\small ywu@stat.ucla.edu,}
{\tt\small\{pingqing,thattg,areganti,premknat\}@amazon.com}
}

\maketitle

\begin{abstract}
Outside-knowledge visual question answering (OK-VQA) requires the agent to comprehend the image, make use of relevant knowledge from the entire web, and digest all the information to answer the question. Most previous works address the problem by first fusing the image and question in the multi-modal space, which is inflexible for further fusion with a vast amount of external knowledge. In this paper, we call for a paradigm shift for the OK-VQA task, which transforms the image into plain text, so that we can enable knowledge passage retrieval, and generative question-answering in the natural language space. This paradigm takes advantage of the sheer volume of gigantic knowledge bases and the richness of pre-trained language models. A \textbf{T}ransform-\textbf{R}etr\textbf{i}eve-\textbf{G}enerate framework (\textbf{TRiG}) framework is proposed\footnote{The code of this work will be made public.}, which can be plug-and-played with alternative image-to-text models and textual knowledge bases. Experimental results show that our \textbf{TRiG} framework outperforms all state-of-the-art supervised methods by at least 11.1\% absolute margin.


\end{abstract}

\section{Introduction}
\label{sec:intro}

The visual question answering (VQA) task is to provide a natural language answer to a natural language question given an image\cite{antol2015vqa}. This task has been well studied in the research communities, and numerous cross-modal methods have achieved state-of-the-art performance \cite{singh2019towards,yu2019deep,jiang2020defense,guo2021bilinear,yang2020trrnet,lu2019vilbert,chen2019uniter,li2020oscar, zhang2021vinvl,li2020unimo}. The knowledge-based visual question answering (KB-VQA) task requires more extensive learning since the questions can be answered only by referring to external general knowledge\cite{shah2019kvqa,wang2015explicit,wang2017fvqa,lu2018r,shah2019kvqa}. Most KB-VQA datasets come with pre-defined knowledge bases, and each question is annotated with at least one supporting knowledge fact. Moreover, the recently proposed outside-knowledge visual question answering (OK-VQA) task is the most open in the sense that any external knowledge can be used to answer the questions.

\begin{figure}[t!]
    \centering
    \includegraphics[width=\linewidth]{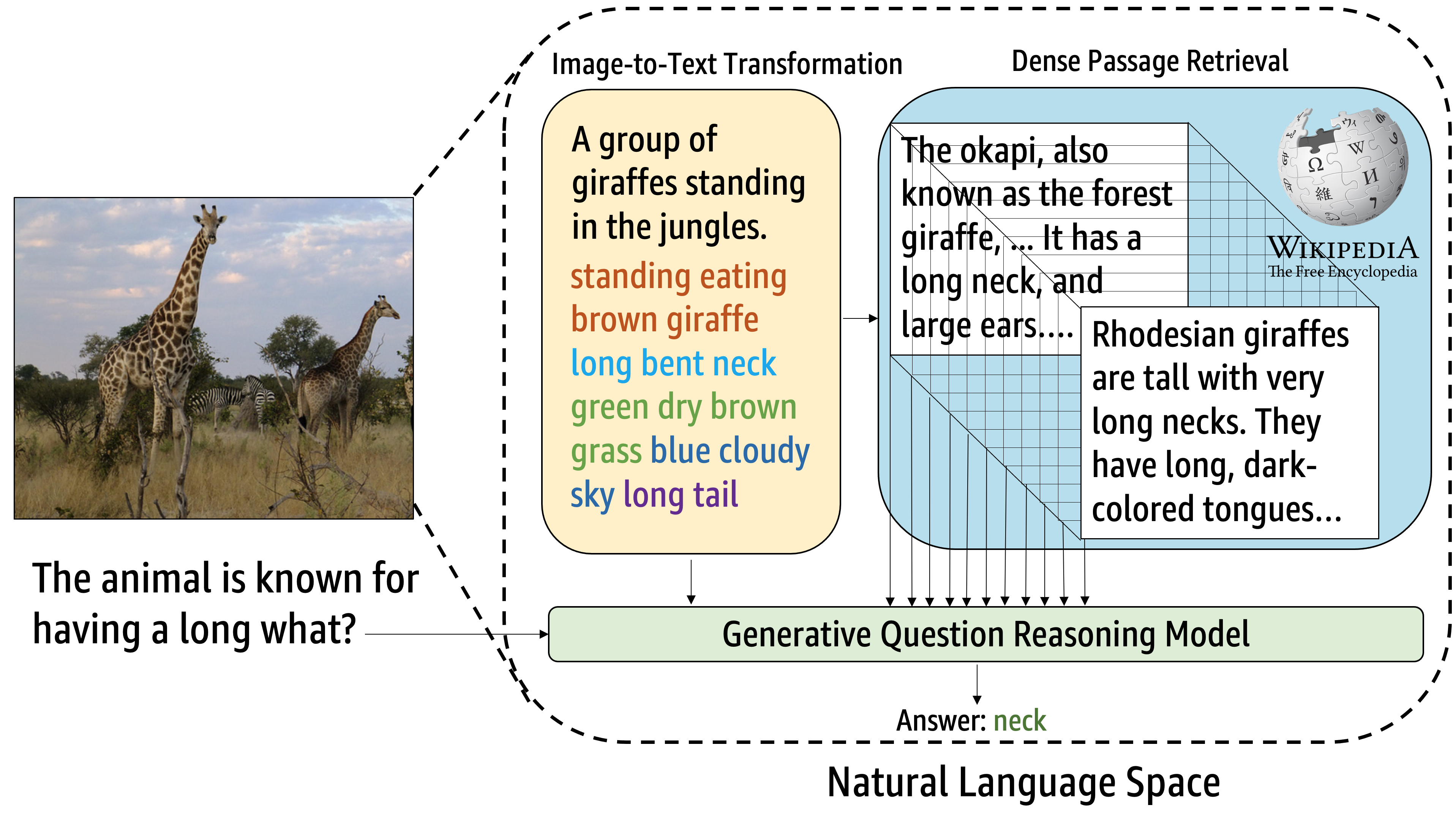}
    \caption{An intuitive example of our TRiG framework on the OK-VQA problem. Our Framework transform all information into language space and performs retrieved-based question answering through generative language models.
    }
    \label{fig:overview}
\end{figure}

Consider the example in \autoref{fig:overview}. As a human, one needs to first identify objects like \textit{giraffes} and \textit{trees} in the image, and associate the \textit{giraffes} to the word \textit{animal} in the question. Second, the human needs to apply his/her acquired commonsense knowledge about \textit{giraffe}'s characteristics and answer the question that \textit{giraffe} is known for having a \textit{long neck}. For machine learning models to solve the same problem, there are several unique challenges. First, in order to answer such a question, one has to align the image, the question, and the vast amount of knowledge passages into one common space. One solution is to first fuse the image and question information in the multi-modal space with pre-trained vision-language models, and then inject knowledge into the multi-modal space. Most previous work on OK-VQA follow this paradigm, including directly injecting the knowledge embeddings \cite{garderes2020conceptbert,shevchenko2021reasoning} and fusing the output of a vision-language model with the knowledge graph through graph convolutional network\cite{marino2021krisp}. However, this paradigm is at the cost of squeezing the rich representation of the textual knowledge, in the magnitude of hundreds of millions, into a much smaller multi-modal space. Comparing to knowledge corpus such as BookCorpus (800M words) and English Wikipedia (2,500M words), multi-modal pretraining datasets are much smaller such as Visual Genome with 0.01 million images and less than 2 million question-answer pairs\cite{krishna2017visual}, which leads to less knowledge. Therefore, we argue that it is possible to transform everything into the language space first, and then take advantage of the tremendous amount of textual knowledge for question answering. Although this seems counter-intuitive, our work proves its advantage. In this paradigm, the challenge is to be able to transform the image into language with minimum information loss. In order to tackle this, we propose three-level image-to-text transformations which significantly outperform baselines that use only captions or object labels. 

The second challenge of the OK-VQA task is how to effectively retrieve the most relevant knowledge passages from gigantic knowledge bases. Previous work has explored various retrieval methods such as term-based BM25\cite{luo2021weakly}, and network-based ranking \cite{luo2021weakly,wu2021multi}. In the OK-VQA dataset, this task is problematic in that there is no ground-truth knowledge annotation for each question. The retrieval has to rely on either transfer learning from similar knowledge-retrieval tasks or weak supervision from pseudo signals such as whether the passage contains the answer tokens\cite{qu2021passage}. Our preliminary study finds that there is no guarantee that a passage containing the ground-truth answer will essentially relate to the question or help the answer prediction. Such signals are very weak and may introduce more noise than useful information into the retrieval model. Instead, we adopt the state-of-the-art dense passage retrieval model (DPR) \cite{karpukhin2020dense} that is pre-trained on large question-answering dataset \textit{Natural}  \textit{Questions} (NQ) \cite{kwiatkowski2019natural} as our knowledge retriever, which is shown to outperform the BM25 method in terms of retrieval coverage rate.

The third challenge of the OK-VQA task is to consolidate all the multi-source input, namely the question, visual context, and the retrieved knowledge passages, to predict answers. Since now everything is in the language space, the problem can be formulated as a multi-passage question answering problem. More specifically, the model needs to not only rank the retrieved passages but also predict an answer according to the ranked passages. Most existing work utilizes extractive methods to predict the answer span in the passage \cite{rajpurkar2016squad,chen2017reading,clark2017simple,rajpurkar2018know,wang2019multi,lee2019latent,yang2019end}. This is not applicable in the OK-VQA dataset because there is neither annotation of ground-truth passage nor answer span in any passage. Instead, we use the generative question answering model \cite{izacard2020leveraging} to avoid the defect in span prediction. Furthermore, we use beam-search for robust answer generation. Lastly, since the question-answering model is the last stage in the entire framework, any information distortion or loss in the image-to-text transformation and knowledge retrieval would propagate to the final question answering model. Therefore, it is important for the final question answering model to be more transparent and interpretable to diagnose the root cause of errors. We use cross-attention scores from the decoder of the generative model to rank and highlight the top supporting knowledge passages, which helps to interpret the results of the model.

To bridge the above-mentioned research gaps, we propose the \textbf{T}ransform-\textbf{R}etr\textbf{i}eve-\textbf{G}enerate (TRiG) framework for the OK-VQA task. At the high level, the framework aligns all the information (image, question, and knowledge) into the language space in order to take advantage of the rich semantics of textual knowledge. The framework starts with three-level image-to-text transformations, followed by dense passage retrieval to retrieve the most relevant knowledge passages. Further, the TRiG aggregates the information from all passages and generates an answer that is relatively easy to interpret. Our contributions are as follows:

\begin{itemize}
    \item
    We propose a new paradigm shift for the OK-VQA task, from aligning all the information in the multi-modal space, to first transforming an image into plain text and performing knowledge retrieval and question answering all in language space. 
    
    \item
   We propose a robust framework \textbf{T}ransform-\textbf{R}etr\textbf{i}eve-\textbf{G}enerate (TRiG), that achieves new state-of-the-art performance on the OK-VQA dataset and leading other supervised methods by 11.1\%. 
\end{itemize}


\section{Related Work}
\label{sec:related_work}
\paragraph{Visual Question Answering (VQA)}
The conventional visual question answering (VQA) task aims to answer questions pertaining to a given image. Multiple VQA datasets have been proposed, such as Visual Genome QA\cite{krishna2016visual} VQA \cite{antol2015vqa}, GQA\cite{hudson2019gqa}, CLEVR\cite{johnson2017clevr}, MovieQA\cite{tapaswi2016movieqa} and so on. Many works have shown state-of-the-art performance on VQA tasks, including task-specific VQA models with various cross-modality fusion mechanisms \cite{singh2019towards,yu2019deep,kim2018bilinear,yu2018beyond,jiang2020defense,guo2021bilinear,yang2020trrnet} and joint vision-language models that are pretrained on large-scale vision-language corpus and finetuned on VQA tasks \cite{lu2019vilbert,chen2019uniter,li2020oscar, zhang2021vinvl,li2020unimo,tan2019lxmert,gan2020large}. Please note that the conventional VQA task does not require external knowledge by definition, although studies show some VQA questions may require commonsense knowledge to answer correctly \cite{antol2015vqa}.

\begin{figure*}[h!]
    \centering
    \includegraphics[width=0.95\linewidth]{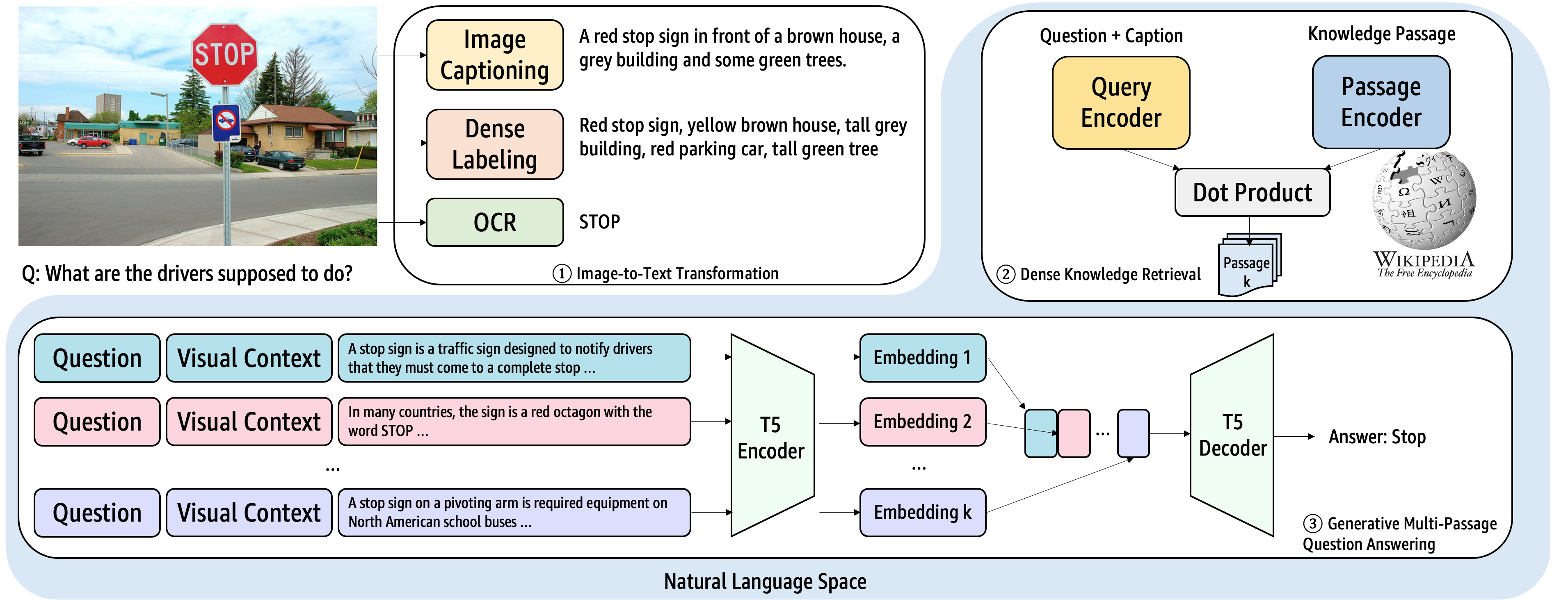}
    \caption{The overview of our \textbf{TR}i\textbf{G} framework. (1) \textbf{T}: Our TRiG framework transforms all visual information into natural language space on three-levels: image-level captioning, object-level dense labeling and text OCR. (2) \textbf{R}: Our dense knowledge retriever retrieve top-k knowledge passages from Wikipedia that are relevant to the query. (3) \textbf{G}: Our generative question answering model encode all question-context-knowledge tuples and fuses the output to generate a final answer. }
    \label{fig:qa}
\end{figure*}

\paragraph{Outside Knowledge-Based VQA (OK-VQA)} 
Beyond the above paradigm, knowledge-based visual question answering (KB-VQA) is proposed where a visual question cannot be answered without external knowledge. Several knowledge-based VQA datasets are proposed, each providing its own knowledge bases and ground-truth supporting fact \cite{shah2019kvqa,wang2017fvqa,lu2018r}. More recently, the dataset outside-knowledge visual question answering (OK-VQA)\cite{marino2019ok} is proposed where the usage of outside knowledge is open to the entire web. Most existing work for OK-VQA rely on the pre-trained vision-language models as a major workhorse for question answering\cite{garderes2020conceptbert,shevchenko2021reasoning,marino2021krisp,wu2021multi,luo2021weakly,yang2021empirical}. In \cite{garderes2020conceptbert,shevchenko2021reasoning}, learned knowledge embeddings are injected into vision-language models to perform knowledge-aware question answering. Other work uses vision-language models as a knowledge-free VQA model first and later adjusts the predicted answers by fusion with knowledge graphs \cite{marino2021krisp} or answer validation with knowledge text \cite{wu2021multi}. Some also propose to directly learn vision-language representation for dense knowledge retrieval \cite{luo2021weakly}. Different from the above, one recent work proposes to first convert the image into text caption and tags and then perform prompt-based QA on GPT-3 model purely in the language space \cite{yang2021empirical}. However, the accessibility to this super-large-scale pre-trained language model is restricted, and it is challenging to interpret the QA result from the generative GPT-3 model. 

\paragraph{Open-Domain Question Answering in NLP}
Open-domain question answering (Open-Domain QA) has been popular in the NLP community in recent years. The task is to answer a question with external knowledge bases without any given context paragraphs\cite{rajpurkar2018know}. There are mainly two streams of approaches, namely knowledge graph-based question answering \cite{sun2018open,lin2019kagnet,wang2019improving,feng2020scalable,lv2020graph,yasunaga2021qa} and knowledge retrieval-based question answering\cite{rajpurkar2016squad,chen2017reading,clark2017simple,rajpurkar2018know,wang2019multi,lee2019latent,yang2019end,izacard2020leveraging}. For retrieval-based methods, both elastic-search such as BM25\cite{robertson2009probabilistic} and semantic search such as Dense Passage Retrieval (DPR)\cite{karpukhin2020dense} are utilized to retrieve most relevant knowledge snippets from knowledge bases. For question answering, most existing work adopt extractive methods to predict the span of an answer in knowledge snippets \cite{rajpurkar2016squad,chen2017reading,clark2017simple,rajpurkar2018know,wang2019multi,lee2019latent,yang2019end}. One most recent work proposes to use generative language models for knowledge-based QA, which achieves state-of-the-art performances \cite{izacard2020leveraging}.

\section{Methodology}
\label{sec:method}
In this section, we introduce the details of our \textbf{T}ransform-\textbf{R}etr\textbf{i}eve-\textbf{G}enerate (\textbf{TRiG}) framework. Shown in \autoref{fig:qa}, our framework contains three stages: (i) image-to-text transformation, (ii) knowledge passage retrieval, (iii) multi-passages open-domain question answer generation.

\subsection{Image-to-Text Transformation}
\label{sec:method:i2t}
Contrary to existing work, we first transform the image into text and then perform all downstream tasks in the language space. In order to minimize the information loss in the process of transforming the image into plain text, three-levels of transformations are performed (\autoref{eq:image_to_text}). First, image-level information is transformed to caption text with a state-of-the-art image captioning model\cite{li2020oscar}. Second, object-level information is translated to object and attribute labels\cite{anderson2018bottom,han2021image}. Lastly, according to \cite{jain2021select}, some VQA questions can only be answered with optical character recognition (OCR). We use an off-the-shelf OCR model to detect all possible texts in the images \footnote{https://github.com/JaidedAI/EasyOCR}.

We denote $C_i$, $L_i$, and $O_i$ as the generated caption text,  attribute and object text, and OCR text from image $I_i$ respectively. In the rest of the paper, we will denote the visual context $v_i = (C_i, L_i, O_i)$ for the corresponding image $I_i$. Please note that our proposed framework does not necessitate the use of the above-mentioned image-to-text transformation models only. One could choose to plug-and-play alternative methods into the framework.

\begin{equation}
    \begin{aligned}
        C_i &= (w^c_0,\dots,w^c_j) \leftarrow f_{Image Captioning}(I_i) \\
        L_i &= \{(w^{attr}_0,w^{obj}_0),\dots,(w^{attr}_n,w^{obj}_m)\} \leftarrow f_{tagging}(I_i) \\
        O_i &= \{w^{ocr}_0,\dots,w^{ocr}_k\} \leftarrow f_{ocr}(I_i)
    \end{aligned}
    \label{eq:image_to_text}
\end{equation}



\subsection{Knowledge Passage Retrieval}
\label{sec:method:dpr}
After the image is transformed into plain-text representation, we use the text representation as the query to retrieve knowledge passages in the natural language space. In this paper, we use the Wikipedia dump as the knowledge base, which contains over 21 million Wikipedia passages \cite{lee2019latent}. We ensure that our framework is designed to be generic enough to support other textual KBs such as GenericsKB\cite{bhakthavatsalam2020genericskb} or the surface forms of graph knowledge bases such as ConceptNet\cite{speer2017conceptnet}.

More specifically, given a textual query $q_i$ of an image $I_i$ and a knowledge base $\mathbb{K} = \{p_j\}$ where each $p_j$ is a knowledge passage, the task is to retrieve top $k$ knowledge passages $P_k = [p_1,p_2,\dots,p_k]$ from $\mathbb{K}$ that are most relevant to the 
query $q_i$, where $k \ll |\mathbb{K}|$. In this paper, we empirically use the query $q_i = (Q_i, C_i)$, where $Q_i$ is the original question and $C_i$ is the the generated caption of corresponding image $I_i$. 




We use dense passage retrieval (DPR) to retrieve the knowledge passages \cite{karpukhin2020dense}. DPR encodes both query and passage with Bert layers that could better capture the semantic similarity between them than term-based retrieval methods such as TF*IDF and BM25\cite{karpukhin2020dense}. First, the query $q_i$ and a passage $p_k$ are encoded with two independent pre-trained BERT encoders \cite{devlin2018bert}. We take the embedding of the [CLS] token $\mathbf{x}_{q_i}$ and $\mathbf{x}_{p_i}$ in the BERT to represent $q_i$ and $p_k$ respectively. Second, a similarity scores $\mathrm{sim}(q_i, p_k)$ is calculated by taking the dot product of the two encoded dense vectors of the query $q_i$ and a passage $p_k$. 

\begin{equation}
    \mathbf{x}_{q_i} = E_Q(q_i),\mathbf{x}_{p_i} = E_P(p_k)
    \label{eq:encoder}
\end{equation}

\begin{equation}
\mathrm{sim}(q_i, p_k) = \mathbf{x}^T_{q_i}\cdot\mathbf{x}_{p_k}
    \label{eq:dpr}
\end{equation}




Because of the tremendous amount of passages in the Wikipedia knowledge base, it is time-consuming to retrieve the top $k$ passages for each query from the knowledge base with over 21 million passages. We leverage an open-sourced indexing engine FAISS\cite{JDH17}, an extremely efficient library to speed up the clustering and indexing of large number of dense vectors. Given a query $q_i$, the dense passage retrieval module will return $k$ passages $P_k = [p_{1},p_{2},\dots,p_{k}]$ from the entire knowledge base $\mathbb{K}$ where $\mathrm{sim}(q_i,p_{1}) >\mathrm{sim}(q_i,p_{2})>\dots>\mathrm{sim}(q_i,p_{k})$ and $k \ll |\mathbb{K}|$. The retrieved passages $P_k$ will be later used for downstream question-answering. 


\subsection{Generative Multi-Passages QA}
\label{sec:method:fid}
After aligning the visual information, the question, and the external knowledge into the language space, we introduce our generative question-answering module. Our design of the model takes the following into consideration. First, although previous work on joint vision-language models formulates the task as an answer classification task \cite{speer2017conceptnet,marino2021krisp,wu2021multi}, our preliminary studies show that language models seem to be less flexible in classifying text into such high-dimensional answer space (over 100k) given a relatively small dataset. Second, although most previous language QA models follow a span-based answer prediction paradigm \cite{rajpurkar2018know,wang2019multi,lee2019latent,yang2019end}, it is impractical in our open-domain setting since there is no ground-truth supporting fact in our task, let alone the ground-truth answer span for prediction. On the other hand, recent work shows that a generative encoder-decoder network can achieve state-of-the-art performance on multiple open-domain QA datasets \cite{raffel2019exploring}, and it avoids span prediction and directly generates a free-form answer. 

To achieve this goal, we use a transformer-like encoder-decoder model T5 as the backbone of our generative question answering module\cite{2020t5}. It is impractical to include all top-$k$ passages in one T5 model. We use T5 model to encode each $(question, visual\ context, knowledge)$ tuple independently and then fuse the $k$ encoded representations to decode an answer following the idea in \cite{izacard2020leveraging}. 

\paragraph{Multi-Passages Question Answer Generation} First, we feed the concatenated sequence of ($Q_i,v_i,p_{i,k}$) into a self-attentive encoder to get per-position hidden embeddings $ \mathbf{z}^{Q_{i,k}}$, where $q_i$ is the question, $v_i$ is the visual context text and $p_i$ is one passage respectively.
\begin{equation}
    \begin{aligned}
    \mathbf{z}^{Q_{i,k}} &= E_{SelfAttn}(Q_i, v_i, p_{i,k}) \\
                &= (z_0,\dots,z_L)
    \end{aligned}
    \label{eq:fid_encode}
\end{equation}
where $z_i$ is the hidden embedding of the $i$-th token in the sequence, $\mathbf{z}^{Q_{i,k}} \in \mathbb{R}^{1\times L \times h}$ is the hidden representation of the sequence, $L = |(Q_i,v_i,p_{i,k})|$ is the length of the sequence and $h$ is the size of the hidden embedding.

Subsequently, we perform the same encoding operation on all $k$ passages to derive $k$ hidden representations:



\begin{equation}
    \begin{aligned}
    \mathbf{z}^{Q_i} = (\mathbf{z}^{Q_{i,1}},\dots,\mathbf{z}^{Q_{i,k}})
    \end{aligned}
    \label{eq:fid_fuse}
\end{equation}
where we concatenate the $k$ hidden embeddings to $\mathbf{z}^{Q_i} \in \mathbb{R}^{(k \cdot L) \times h}$. This operation is to fuse all the information from different question-context-passage interactions together in order to generate better answers. Then, we feed the concatenated hidden representation $\mathbf{z}^{Q_{i}}$ into a stacked self-attentive decoder to predict per-position word distribution over the vocabulary space $|V|$:

\begin{equation}
    \begin{aligned}
    P (a_1),\dots,P (a_l) = \sigma (D_{SelfAttn}(\mathbf{z}^{Q_{i}})) 
    \end{aligned}
    \label{eq:ans_pred}
\end{equation}
where $\sigma$ is a non-linear function such as softmax, $l$ is the length of the answer, and $Q_i \in \mathbb{R}^{|V|}$ is the word distribution over the vocabulary of size $|V|$. Finally, we use teacher-enforcing to train the entire model with auto-regressive cross-entropy loss:
\vspace{-8pt}
\begin{equation}
    \begin{aligned}
    L_{ans} = - \frac{1}{N \cdot l \cdot |V|} \cdot \sum_{i=1}^{N} \sum_{j=1}^{l} \sum_{w=1}^{|V|} y_{i,j,w} \cdot \log(p(a_{i,j,w})) 
    \end{aligned}
    \label{eq:loss}
\end{equation}

\vspace{-12pt}
\paragraph{Inference of the Multi-Passage Generative Model} 
During training, teacher-enforcing is used to train the encoder-decoder model auto-regressively. During inference time, the answer tokens are generated iteratively by feeding the previous token $a_{t-1}$ to the input of the next token $a_t$. We apply both greedy-decode and beam-search for the answer decoding. In greedy-decode, the best answer token is always selected with the highest probability at each decoding step. In beam search, a beam of size $m$ is maintained during decoding, and $m$ answer candidates are generated with ranked scores. We also take ensembles of the 6 TRiG models trained on different splits of the top-100 passages, where the best answer is selected by ranking the model answers with average log probability of all the generated tokens of the predicted answer: $a^* = \argmax_n \{\frac{1}{l}\sum_j^l \ln P(a_{n, j})\}$ and $n$ is the number of ensembles.



\section{Experiments}
\label{sec:experiments}
In this section, we describe the implementation details of our method and report the experimental results.

\subsection{Implementation Details}
\label{sec:ecperiments:implemenation}
\paragraph{OK-VQA Dataset} We use the OK-VQA dataset in this research work (version v1.1\footnote{https://okvqa.allenai.org/download.html}, license CC-BY 4.0\footnote{http://creativecommons.org/licenses/by/4.0/}). It is one of the most challenging visual question answering datasets that is open to all external knowledge usage\cite{marino2019ok}. The dataset contains 14,055 visual questions over 14,031 images from MSCOCO \cite{lin2014microsoft}. The dataset split is 9,009 for training and 5,046 for testing. Each entry contains an image, a question, and $10$ ground-truth answers annotated by human annotators.

\paragraph{Dense Passage Retrieval} We use BERT-base encoders, $E_Q$ and $E_P$, in the retrieval module and initialize them with the checkpoints pre-trained on the NQ dataset\cite{kwiatkowski2019natural}. Due to the extremely large size of the Wikipedia knowledge base, we choose the HNSW indexing algorithm instead of flat indexing for a much faster speed of queries with acceptable accuracy trade-off. For more details, please refer to the implementation of \cite{JDH17}. Each query is composed of the question $Q_i$ and the corresponding caption $C_i$. The number of retrieved passages $k=100$ for the best possible QA performance. 

\paragraph{Generative Multi-Passages QA} We use a  transformer-based\cite{vaswani2017attention} encoder-decoder T5-large\cite{2020t5} model as the backbone. By default, the embedding size of the encoder is $768$. The maximum length of the input tokens is restricted to be $300$. Padding to the maximum length is applied for multiple questions batch training. Because the training of the generative model with 100 passages is memory-intensive, the batch size is set to be $1$ for each GPU. To optimize the QA model, we apply the following techniques: (i) AdamW as the optimizer with a linearly scheduled learning rate starting from $1e-4$; (ii) Warm-up of 2000 steps as the learning rate scheduler. We train the multi-passages QA model for 20000 optimization steps on an 8xA100 GPU cluster for 12 hours. During inference, both greedy-decode and beam-search are applied to get the best answers. Before evaluation, a normalization step is performed on the generated answers, including lower-casing and removing articles, punctuation, and duplicated white space. 



\subsection{Empirical Results on OK-VQA}
\label{sec:experiments:results}
\subsubsection{Performance of Knowledge Retrieval} 

To evaluate the performance of the knowledge passage retrieval module, we consider a question that has a $hit$ in its retrieved knowledge passages if at least one of its ground-truth answers appears in the retrieved passages. Then the $hit@k$ is defined as the percentage of questions in the entire dataset who get a $hit$ in their top $k$ retrieved knowledge passages.


\begin{table}[ht!]
    \centering
    \begin{tabular}{l c c}
        \hline
                            & OK-VQA Train & OK-VQA Test \\
        \hline              
        Top-K               & hit@k         & hit@k      \\
        \hline
        Top-5               & 42.72\%       & 45.83\%    \\
        Top-10              & 54.66\%       & 57.88\%    \\
        Top-20              & 68.76\%       & 72.11\%    \\
        Top-50              & 72.27\%       & 80.49\%    \\
        Top-100             & 83.76\%       & 86.56\%    \\
        \hline
    \end{tabular}
    \caption{Hit@k of the dense passage retrieval (k = the number of retrieved knowledge passages).}
    \label{tbl:wiki_retrieval}
\end{table}

From \autoref{tbl:wiki_retrieval}, we can observe that the answer retrieval rate $hit@k$ increases along with the number of passages $k$ from 42.7\% to 83.7\% as $k$ increases from 5 to 100. 
A larger $k$ increases the probability that each question has access to at least one relevant knowledge passage during inference. We experiment with different $k$ for the downstream QA model, which will be discussed in \autoref{sec:experiments:ablation}.


\subsubsection{Performance of the Generative QA Model}
\paragraph{Exact Match and VQA Score} The OK-VQA dataset has 10 annotated answers for each question, and we consider both Exact Match and VQA Score as metrics to evaluate the generative QA model. The Exact Match (EM) is defined as the percentage of questions whose predicted answer exactly matches any of the 10 annotated answers. EM metric considers every answer as equally ground-truth the same. On the other hand, VQA score defines a voting mechanism so that each annotated answer $a_i$ is assigned a score $s_i$ between 0 and 1\cite{antol2015vqa}.




A generated answer $\hat{a}_i$ would get $s_i$ score if it matches the annotated $a_i$. The VQA metric is an average of the weighted scores over the entire test set. Arguably, the voting mechanism of the VQA score may promote some ground-truth answers over others based on the annotators' consensus subjectively.


\paragraph{Comparison with Supervised-Learning SOTAs}
The performance of our proposed TRiG framework with state-of-the-art models is reported in \autoref{tbl:supervised_sota}. Please note that all the models in comparison are supervised-learning models. Several observations can be made from the table. First, most previous methods utilize the vision-language model as the backbone for question answering and then integrate it with external knowledge. Some represent the knowledge in the form of graph (KRISP\cite{marino2021krisp}, ConceptBert\cite{garderes2020conceptbert}, RVLESK\cite{shevchenko2021reasoning}) while others fuse the output of the vision-language model with textual knowledge representation (MAVEx\cite{wu2021multi}) or implicit knowledge from a language QA mdoel\cite{salaberria2021image}. Second, a concurrent work, VRR\cite{luo2021weakly}, transforms the image into caption text and performs span-based question answering on a trimmed knowledge base using Google search engine. Last and most importantly, all of the above methods achieve very similar VQA scores between 38.60 and 39.4, despite usage of diverse sources of knowledge bases such as ConceptNet\cite{wu2021multi,marino2021krisp,garderes2020conceptbert,shevchenko2021reasoning}, Google Image\cite{wu2021multi}, Google Web Search\cite{luo2021weakly} and Wikipiedia\cite{wu2021multi} and pretraining on other datasets such as VQA\cite{garderes2020conceptbert,marino2021krisp,shevchenko2021reasoning,wu2021multi} and Visual Genome\cite{shevchenko2021reasoning}.


Our proposed TRiG framework significantly outperforms all state-of-the-art supervised-learning methods with at least a 11.1\% margin. Our TRiG framework differs from the existing methods as (i) instead of aligning representation of the vision-language QA model with external knowledge in the multimodal space, TRiG transforms the image into text information as accurately as possible and aligns all the information of the image, question, and knowledge in language space; (ii) the generative QA model in TRiG is not pre-trained on other multimodal datasets, which helps the model to start learning to reason over external knowledge, rather than inducing data bias from other multimodal datasets.

\begin{figure*}[h!]
    \centering
    \includegraphics[width=1.0\linewidth]{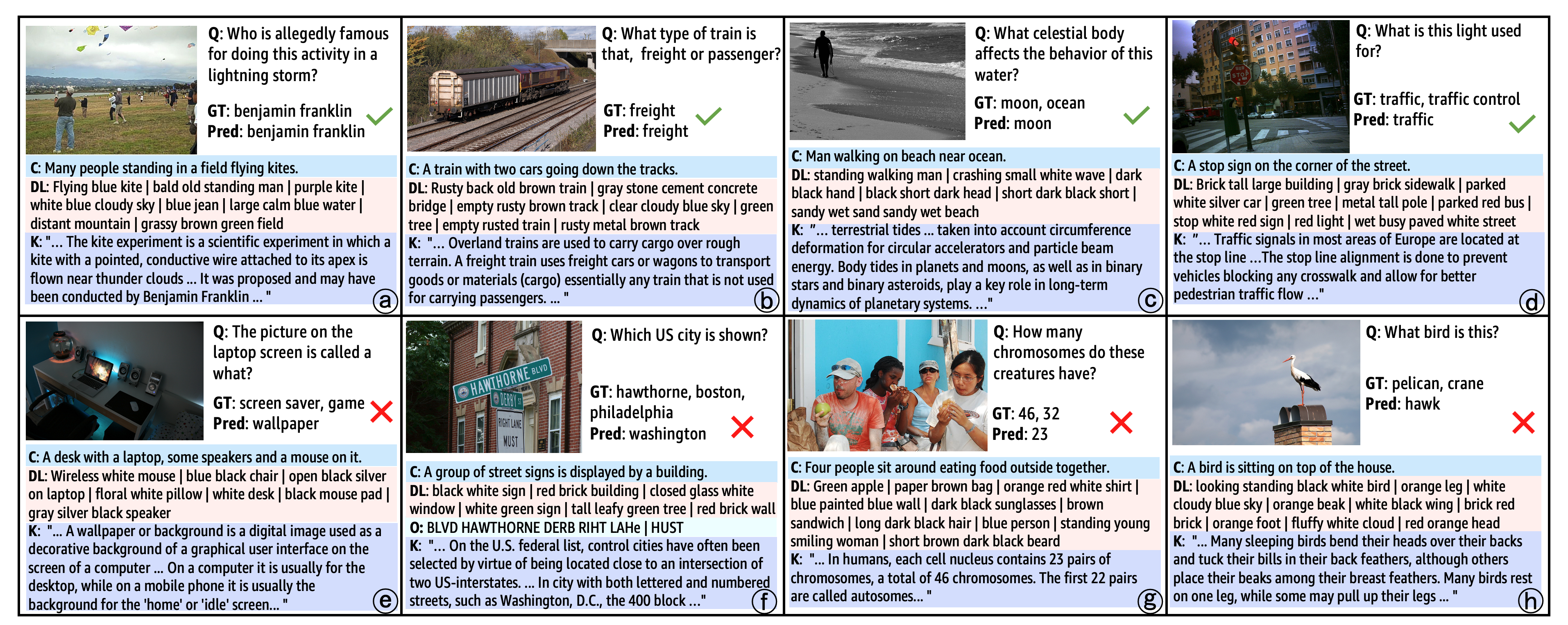}
    \caption{Examples of our TRiG model prediction together with the supporting passage. \textbf{Top}: four examples where TRiG model makes correct predictions. \textbf{Bottom}: four examples where TRiG model makes incorrect predictions. In each example: \textbf{Q}: question, \textbf{GT}: ground-truth answers, \textbf{Pred}: predicted answer, \textbf{C}: image caption, \textbf{DL}: dense labels, \textbf{O}: OCR text, \textbf{K}: top-1 supporting knowledge passage.}
    \label{fig:case_examples}
\end{figure*}

We would like to also highlight the Exact Match (EM) score of our TRiG models, which are higher than the VQA scores. As in \autoref{fig:case_examples}, we observe that sometimes the generative QA model predicts a reasonable answer but is not credited with the highest VQA score or not even any score according to annotators' voting. 


\begin{table}[t!]
    \centering
    \begin{tabular}{l c c}
        \hline
        \textbf{Model}                                  & EM        & VQA Score \\
        \hline
        \textbf{SOTA Methods}                           &           &           \\
        \hline
        KRISP\cite{marino2021krisp}                     &           & 32.31     \\
        ConceptBert\cite{garderes2020conceptbert}       &           & 33.66     \\
        CBM\cite{salaberria2021image}                   &           & 38.60     \\
        KRISP w/ VQA2.0 pretrained                      &           & 38.70     \\
        MAVEx\cite{wu2021multi}                         &           & 38.70     \\
        RVLESK\cite{shevchenko2021reasoning}            &           & 39.04     \\
        Weakly Supervised VRR\cite{luo2021weakly}       &           & 39.20     \\
        MAVEx w/(Ensemble 5) \cite{wu2021multi}         &           & 39.40     \\
        \hline
        \textbf{Ours}                                   &           &           \\
        \hline
        TRiG w/ Q+C+DL+O, G                             & \textbf{53.62\%}  & \textbf{49.24}   \\
        TRiG w/ Q+C+DL+O, BS                            & \textbf{53.59\%}  & \textbf{49.35}   \\
        TRiG w/ Q+C+DL+O, G, $\mathbf{E}^*$             & \textbf{54.73\%}  & \textbf{50.50}   \\
        \hline
    \end{tabular}
    \caption{Comparison of supervised-learning methods on the OK-VQA dataset. In TRiG Model, \textbf{Q}: Question, \textbf{C}: Caption, \textbf{DL}: Dense Labels, \textbf{O}: OCR Text, \textbf{G}: Greedy Decode, \textbf{BS}: Beam-Search, \textbf{$\mathbf{E}^*$}: Ensembles of the 6 TRiG models.}
    \label{tbl:supervised_sota}
\end{table}


\begin{table}[ht!]
    \centering
    \begin{tabular}{l c c}
        \hline
        \textbf{Model}                                                                  & \#Params        & VQA Score \\
        \hline
        \textbf{SOTA Prompt Method}\cite{yang2021empirical}                             &           &           \\
        \hline
        PICa w/16 RP C+T                                                                &        175B   & 43.30     \\
        PICa w/16 SP C+T                                                                &        175B   & 46.50     \\
        PICa w/16 SP C+T, 3$\times\text{E}$                                             &        175B   & 47.70     \\
        PICa w/16 SP C+T, 5$\times\text{E}$                                             &        175B   & 48.00     \\
        \hline
        \textbf{Ours}                                                                   &           &           \\
        \hline
        TRiG w/ Q+C+DL+O, G                                                             & 0.77B      & \textbf{49.24}  \\
        TRiG w/ Q+C+DL+O, BS                                                            & 0.77B     & \textbf{49.35}   \\
        TRiG w/ Q+C+DL+O, G, $\mathbf{E}^*$                                             & 0.77B     & \textbf{50.50}   \\
        \hline
    \end{tabular}
    \caption{Comparison of Proposed TRiG with SOTA Prompt-Based Method on the OK-VQA Dataset. In\cite{yang2021empirical}, \textbf{RP}: Random Prompt, \textbf{SP} Selected Prompt, \textbf{C}: Caption, \textbf{T}: Image-Tagging, \textbf{E}: Prompt Ensemble. In TRiG model, \textbf{Q}: Question, \textbf{C}: Caption, \textbf{DL}: Dense Labels, \textbf{O}: OCR Text, \textbf{G}: Greedy Decode, \textbf{BS}: Beam-Search, \textbf{$\mathbf{E}^*$}: Ensembles of the 6 TRiG models.}
    \label{tbl:gpt3}
\end{table}
\vspace{-12pt}

\paragraph{Comparison with Prompt-Based SOTA}
We also compare our method with one very recent prompt-based method on the OK-VQA problem \cite{yang2021empirical}. By taking advantage of the super large-scale language model GPT-3 \cite{brown2020language}, the proposed prompt-based method (PICa) surpasses all existing supervised methods with sophisticated prompting. As shown in \autoref{tbl:gpt3}, PICa achieves 43.3 VQA score with 16 prompts randomly selected from the training data. By carefully selecting 16 prompts based on the similarity between testing and training questions, PICa further achieves 46.5. With 5 ensembles of 16 prompts, PICa reaches 48.0 VQA score.

Our method (TRiG) outperforms PICa with greedy-decode 49.24, beam-search decoding 49.35 and ensemble 50.50. Both PICa and our method share the same idea of unifying the image, the visual question, and knowledge in language space and then performing question answering with language models. The significant performance gain of both methods (9-11.1\% over SOTA) highlights the potential of this idea -- if the image could be transformed into plain text information faithfully, then one could take advantage of the vast volume of external knowledge in text form and advanced language models pre-trained on rich variations of human natural language to yield better answer prediction.

We would like to also highlight that our method outperforms PICa by a margin of 2.50\%, especially considering the among of parameters (175 billion over 0.77 billion of our model) and accessibility of the GPT-3 model. Moreover, we argue that our prediction results are relatively easier to interpret by selecting supporting knowledge passages, whereas in PICa the explanation is generated by GPT-3 in a black-box manner. We use the averaged cross-attention score of the generative model to select supporting facts\cite{izacard2020distilling}. For concrete examples of such interpretability, please see the examples in \autoref{fig:case_examples}. 




\subsection{Ablation Study}
\label{sec:experiments:ablation}
\paragraph{Variant Visual Context Input} We investigate the empirical differences among the combination of the visual contexts inputs to the generative QA model, namely image caption (C), object label (L and DL), and OCR (O).

\begin{table}[h!]
    \centering
    \begin{tabular}{l c}
        \hline
        Inputs                  & VQA Score   \\
        \hline 
        Question + K + C            & 42.54       \\
        \hline
        Question + K + C + L        & 42.94       \\
        \hline
        Question + K + C + L + O    & 43.53       \\
        \hline
        Question + K + C + DL + O   & \textbf{49.35}       \\
        \hline
    \end{tabular}
    \caption{Ablation Study of the Different Variants of Text Input into the Generative QA Model (\textbf{K}: Knowledge passages, \textbf{C}: Caption, \textbf{L} = Bottom-Up Labels\cite{anderson2018bottom}, \textbf{DL} = Dense Labels, \textbf{O} = OCR Text).}
    \label{tbl:visual_ablation}
\end{table}

\begin{figure}[h!]
    \centering
    \includegraphics[width=\linewidth]{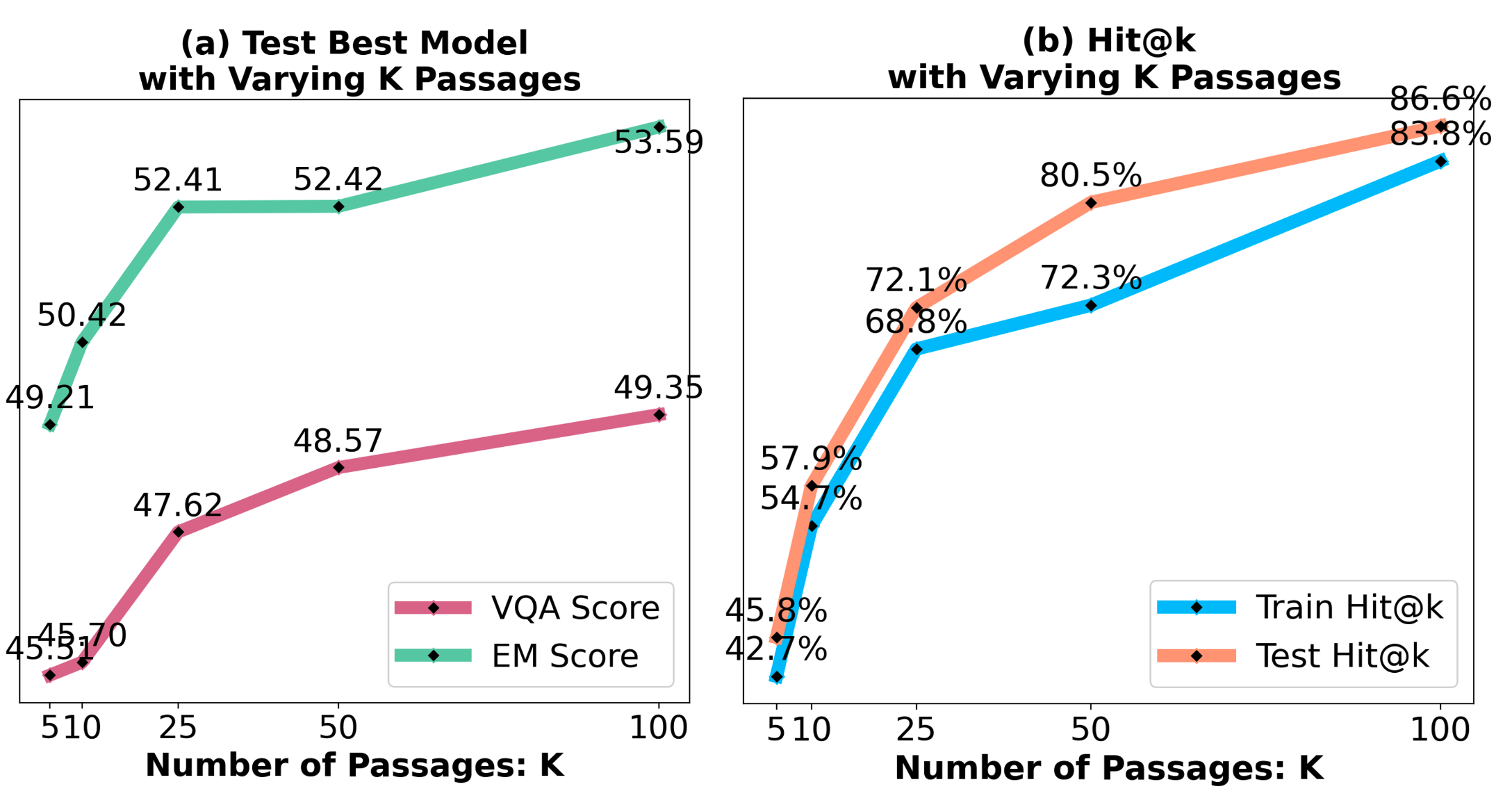}
    \caption{Testing the QA model with varying number of passages.}
    \label{fig:vary_k}
\end{figure}
\vspace{-12pt}

As in \autoref{tbl:visual_ablation}, we find that adding caption (C) to the input yields decent performance (42.5), suggesting that caption conveys basic information of the image. Adding sparse object labels and attributes (L) also helps a little (42.9). By adding OCR, the performance is further improved (43.5), which is in accord with previous findings that some questions in OK-VQA require understanding the text in the image through OCR\cite{jain2021select}. Interestingly, the largest gain is achieved by replacing sparse object and attribute labels with more semantically rich dense object labels (49.4), which again highlights that the faithfulness of image-to-text transformation is a crucial prerequisite for downstream QA in the language space.

\paragraph{Generative Multi-Passages QA with Varying K passages} We also investigate how the generative QA model behaves with a different number of passages $k$. We apply our best model trained on 100 passages and test it with varying $k$ passages. From \autoref{fig:vary_k}-(a), we can see that the testing performance of this model steadily increases along with the growing number of passages $k$. However, the improvement becomes marginal after $k$=25 (47.62 to 49.35), while the coverage \textit{Hit@k} still increases by 15\% as \autoref{fig:vary_k}-(b). This also supports our hypothesis that there may be a long-tail effect of the retrieval. Yet it is difficult to quantify as to which passages are essentially relevant to the question-answering.

\subsection{Discussion}
\begin{figure}[ht!]
    \centering
    \includegraphics[width=\linewidth]{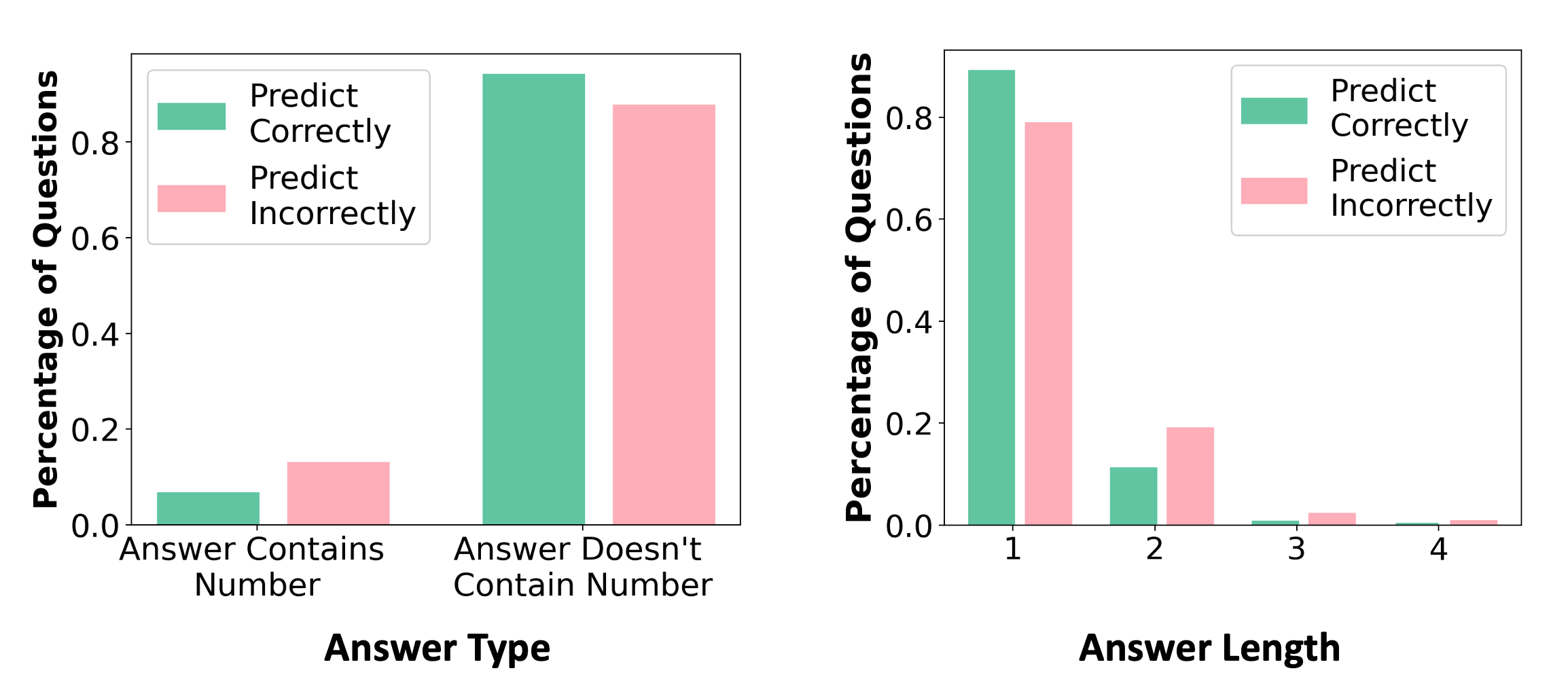}
    \caption{Performance of Generative QA model by different answer types. \textbf{Left}: whether numerical answers are harder to predict. \textbf{Right}: whether longer answers are harder to predict.}
    \label{fig:error_types}
\end{figure}

\paragraph{Error Analysis}
To investigate the behavior of our TRiG model, we conduct error analysis with our best model using greedy-decoded predictions. The quantitative results are illustrated in \autoref{fig:error_types}. We observe that answers with numerical values are harder to predict, where the model could get into a blunt generation (\autoref{fig:error_types}-(a)). Furthermore, as the length of the answer increases, it is harder for the generative model to predict every token in the phrase correctly  (\autoref{fig:error_types}-(b)). 

We also manually reviewed 50 examples where TRiG makes wrong predictions. Among these random examples, 50\% of the errors are due to the information loss during image-to-text transformation, such as in \autoref{fig:case_examples}-(h), where the caption and dense labels failed to characterize the special features of the bird. We also found that 24\% of the error are due to the failure in retrieving highly-relevant passages. The high Hit@k value doesn't guarantee the passages are indeed relevant to the question. Note that some examples failed due to multiple reasons including QA error (22\%) or subjective human annotations (30\%) as in \autoref{fig:case_examples}(g). 

\vspace{-12pt}
\paragraph{Interpretability}
To interpret the visual question-answering models, previous works attempt to supervise the VQA models with visual grounding annotations\cite{zhang2019interpretable,das2017human,das2017human} or neural symbolic network\cite{vedantam2019probabilistic,yi2018neural,hu2017learning}. When it comes to knowledge-based VQA, it is all the more challenging to interpret the model in multimodal space because the knowledge has been transformed into a fused representation and loses its meaning. 

Our TRiG framework alleviates this problem by providing transparent explanations in the language space. In the top row in \autoref{fig:case_examples}, the image-to-text transformations provide sufficient information for both the knowledge retrieval and QA model. Meanwhile, when \autoref{fig:case_examples}(e, f, g, h) make wrong predictions, the QA model is still predicting the answer according to the visual context and retrieved passages.





\paragraph{OK-VQA Evaluation Metrics}
Some researchers\cite{luo2021just} also argue that the VQA score metric is subjective. In one OK-VQA example, a model will achieve 1.0 VQA score for the answer \textbf{\textit{wetsuit}} but only 0.66 score for the answer \textbf{\textit{wet}} \textbf{\textit{suit}}.  In daily language, the usage of any of the semantically-similar answers is subtle and sometimes random. We also look at the top-3 answers of our TRiG model using beam search, and the model achieves significantly higher performance, \ie \textbf{67.4} VQA score and \textbf{71.8\% }EM. We call for better VQA metrics that probably compare two sets of answers instead of comparing only the top one answer or other alternatives such as AAS that automatically expands the ground-truth answer set for better matching\cite{luo2021just}.

\section{Conclusion}
In this paper, we approach the OK-VQA task from a new perspective, where all the visual information is aligned into the language space to take advantage of the comprehensiveness in textual knowledge bases. Moreover, we propose a robust Transform-Retrieve-Generate (TRiG) framework that outperforms state-of-the-art supervised methods by 11.1\%. One can plug-and-play with different image-to-text methods and textual knowledge bases into TRiG for potential further improvement. Our work has limitations that the dense passage retrieval is not optimized for the OK-VQA task, due to the unavailability of ground-truth supporting facts. We consider this as one of our future work, as well as improving the quality of image-to-text transformation. 




\newpage
{\small
\bibliographystyle{ieee}
\bibliography{reference}
}

\clearpage
\newpage
\section{Additional Details for Methodology}
\label{sec:method}
\subsection{Generative Multi-Passages QA Details}

\paragraph{Hyper-parameters}
To better illustrate the implementation of the generative multi-passages QA model, we introduce some key hyper-parameters in \autoref{tbl:hyperparam}.

\begin{table}[h]
    \centering
    \begin{tabular}{l c}
        \hline
        Hyper-Parameter                 & Value     \\
        \hline 
        \hline
        Max Input Length                & 300       \\
        \hline
        Max Decoding Length             & 20        \\
        \hline
        Early Stopping                  & True      \\
        \hline
        Pad to Max Length               & True      \\
        \hline
        Max Number of Beams             & 3         \\
        \hline
        Learning Rate                   & 0.0001    \\
        \hline
        LR Scheduler                    & Linear    \\
        \hline
        Total Optimization Step         & 20000     \\
        \hline
    \end{tabular}
    \caption{Hyper-parameters of the generative multi-passages QA model, not including hyper-parameters for T5 backbone.}
    \label{tbl:hyperparam}
\end{table}

\paragraph{Input Format}
Different from the default input format to the pre-train a T5 model, we use a alternative formatting for the input sequences. We concatenate the question, the visual context and one retrieved Wikipedia knowledge passage as the input sequence, without any special token such as ``\textit{[SEP]}" between them. The question has a prefix ``\textit{question: }" before it. The visual context is the concatenation of image caption, dense labels and OCR text. The knowledge passage consists of a Wikipedia title and a Wikipedia paragraph. The two are concatenated by putting a prefix ``\textit{title: }" and a prefix ``\textit{context: }" before them respectively. 


\paragraph{Vocabulary}
We also want to highlight the effect of the different sizes of QA model vocabulary. As in \autoref{tbl:vocabulary}, we notice a trend that models with larger vocabulary sizes achieve higher performance. In particular, models using the default vocabulary (PICa and TRiG) perform better on OK-VQA dataset. 

\begin{table}[h]
    \centering
    \begin{tabular}{l c c }
        \hline
        Method                          & Size      & VQA Score \\
        \hline 
        \hline
        KRISP w/o VQA2 pre-train        & 2,250      & 32.31     \\
        Weakly Supervised VRR (C)       & 11,060     & 36.78     \\
        RVLESK                          & 14,456     & 39.04     \\
        PICa (5 Ensembles)              & 50,257     & 48.00     \\
        \hline
        \textbf{Ours} (6 Ensembles)     & 32,128     & \textbf{50.50}     \\
        \hline
    \end{tabular}
    \caption{The vocabulary size and performances of different SOTA methods on OKVQA. (C) represents classification. Some numbers may not be public accessible and we only report the numbers directly from the authors.}
    \label{tbl:vocabulary}
\end{table}



\section{Additional Details for Ablation Study}
\label{sec:ablation}
\subsection{Answer Accuracy in Beam-Search}
In the main paper, we argue that the ground-truth answers of an OK-VQA question might be a semantically-similar cluster, such as (\textit{swimsuit}, \textit{bath suit}, \textit{bikini}). This may also hold true for the question answering models, in terms of both classification models (top-k class prediction) and generative models (top-k beam prediction). 

\begin{table}[ht!]
    \centering
    \begin{tabular}{l c c}
        \hline
                            & Exact Match           & VQA  \\
        \hline              
        Top-1               & 53.59\%               & 49.35    \\
        Top-2               & 65.99\%               & 61.61    \\
        Top-3               & \textbf{71.78}\%      & \textbf{67.48}    \\
        \hline
    \end{tabular}
    \caption{Ablation on Different $k$ in Beam-Search Decoding.}
    \label{tbl:top_k_beamsearch}
\end{table}
 
We report the performance of our generative question answering model using top-1/2/3 beam-search decoding. As shown in \autoref{tbl:top_k_beamsearch}, we can find that the both the Exact Match (EM) and VQA score increase as the $k$ of beam-search increase. This suggests that while the top-one answers only achieve 49.35 VQA score, their semantically-similar candidates could reach as high as 67.48 VQA score. Therefore, we call out for new metrics that compare two sets of answers instead of top-one answer versus many ground-truth answers. 

\section{Additional Details for Error Analysis}
To further understand the behavior of our TRiG framework, we conducted several error analysis.

\begin{figure}[h!]
    \centering
    \includegraphics[width=\linewidth]{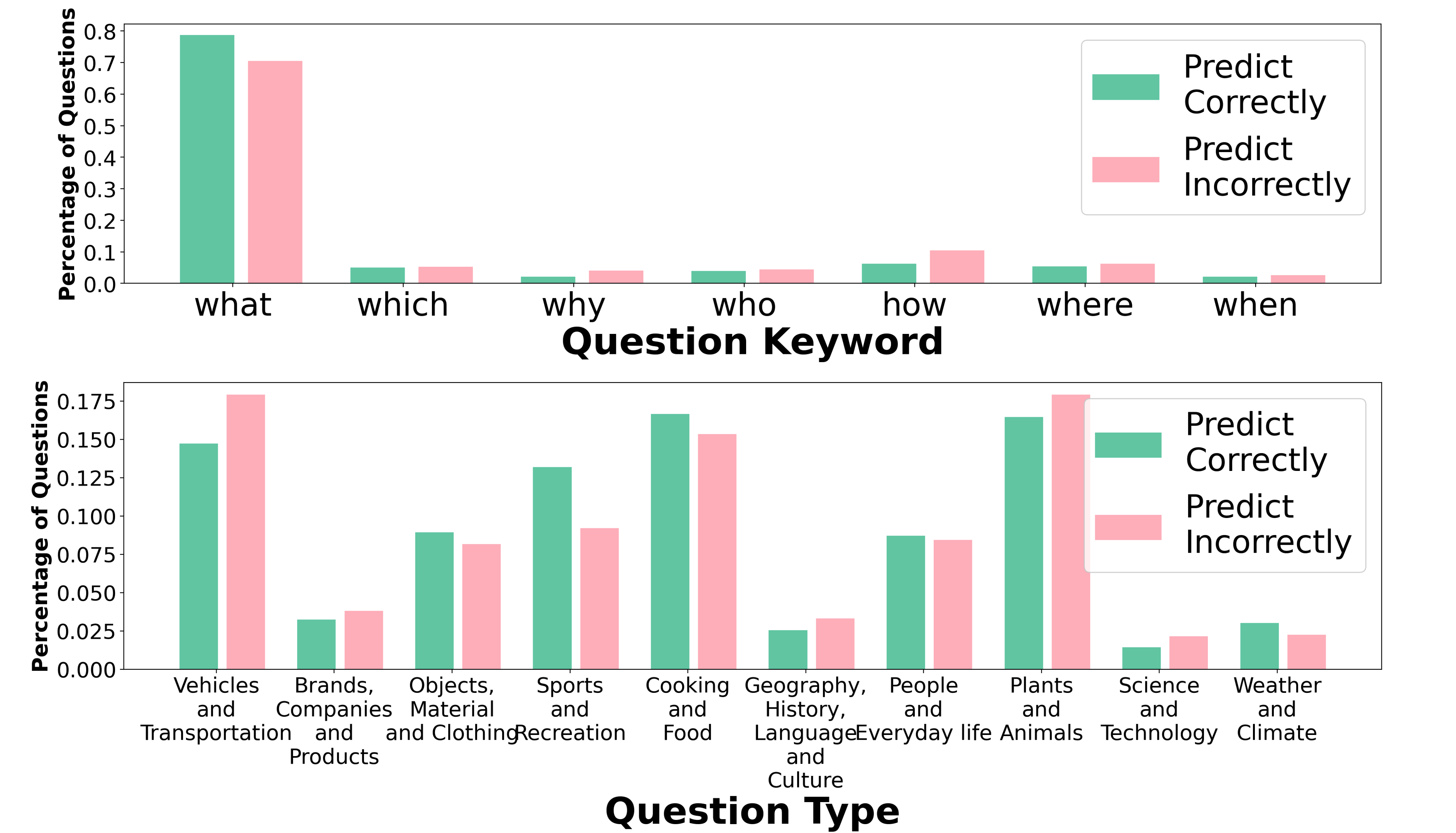}
    \caption{Distribution of correct/incorrect predictions. Top: Distributions of predictions over different question keywords. Bottom: Distribution of predictions over different question types.}
    \label{fig:supp:error}
\end{figure}

 \paragraph{Question Keywords / Types} First, we investigate whether the model is likely to predict correctly over some question keywords than others. As in \autoref{fig:supp:error}-top, we can observe that majority of the questions contain the keyword ``\textit{what}", where our model is more likely to make correct predictions. On the other hand, for questions containing keywords such as ``\textit{how}" and ``\textit{why}", our model is more likely to make mistakes. We hypothesize that the ``\textit{how}" and ``\textit{why}" questions usually entail longer answers, which is harder for the generative model to predict. For example, for the question  \textit{why is this sign here}? (a sign for animal protection), the ground-truth answers are (\textit{protect animal}, \textit{safety}, \textit{don't feed animal}, \textit{direct}).

Second, we report prediction distribution over 10 question types that are available from the OK-VQA dataset. As in \autoref{fig:supp:error}-bottom, we can see that our model is more likely to predict correctly on category of sports and recreation. On the contrary, our model makes more mistakes in Vehicles and Transportation and Plants and Animals.

\begin{figure}[h!]
    \centering
    \includegraphics[width=\linewidth]{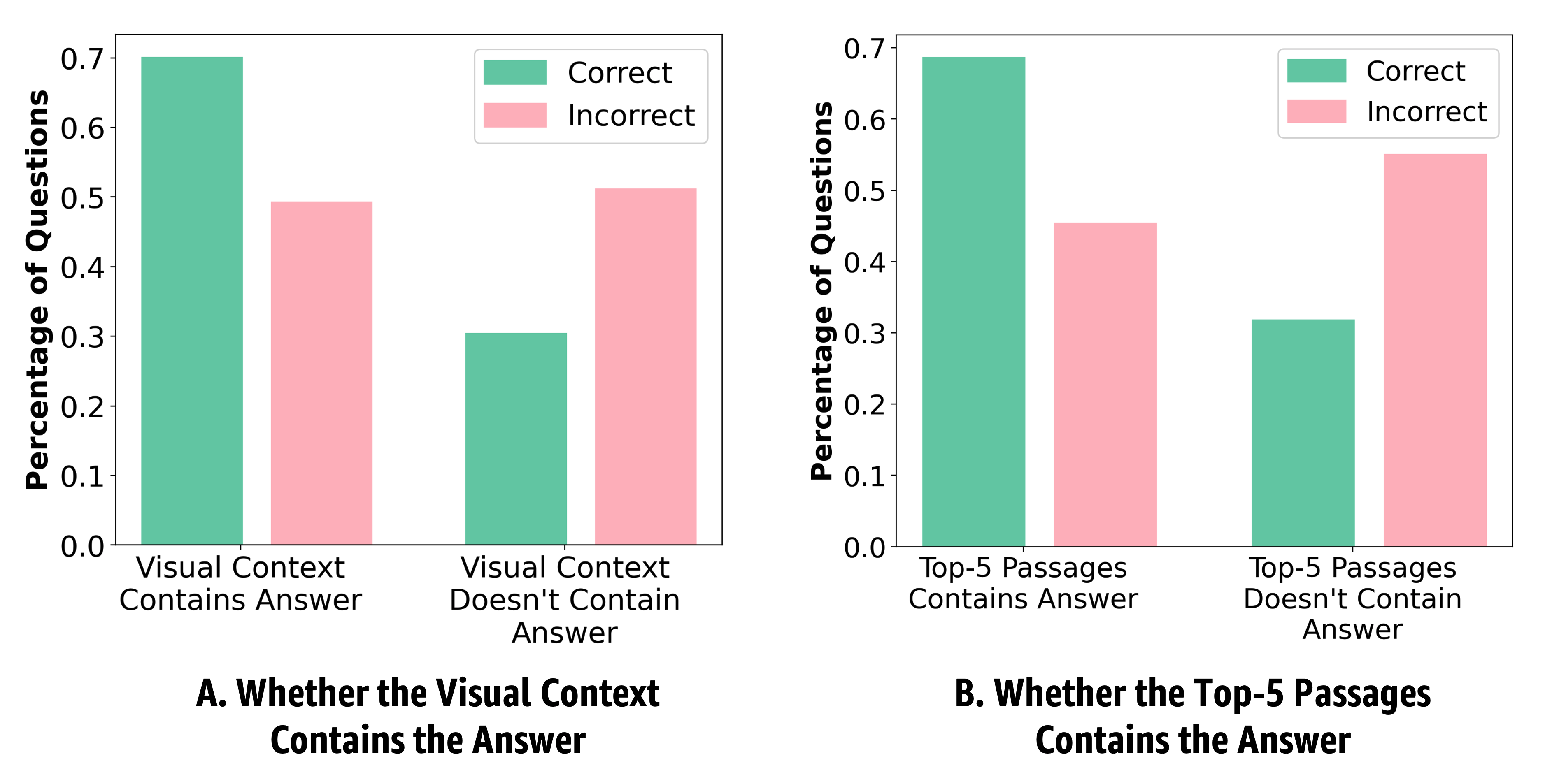}
    \caption{Error Category Break-Down.}
    \label{fig:supp:contain}
\end{figure}

\paragraph{The Impact of Visual Context and Knowledge Passages}  First, we would like to further investigate the effectiveness of the image-to-text transformation module, since it is the first stage in our TRiG framework. Shown in \autoref{fig:supp:contain}-A, we find that if the visual contexts contain the ground-truth answers, the generative question answering model is more likely to generate a correct answer. In contrast, the model makes more mistakes if the visual contexts do not contain the ground-truth answers. 

Second, we also investigate how the retrieved passages impact the generative question answering model. As is illustrated in \autoref{fig:supp:contain}-B, we find that if the top-5 passages that contain the ground-truth answers, our generative question answering model is much more likely to predict correct answers. On the opposite side, if top-5 passages do not contain the ground-truth answers, it is more likely for the QA model to make a wrong prediction. 


\paragraph{Manual Error Review} We also conducted manual eye-balling on 50 random examples where the model has made wrong predictions. We look into each example with all the available information (question, caption, dense labels, OCR text, knowledge passages, ground-truth answers) and attribute each example to one or more error categories. A brief statistics is shown in \autoref{tbl:error_analysis}. Please note that the percentages of error types are not mutually exclusive because some wrong cases may fall in multiple categories.


\begin{table}[ht!]
    \centering
    \begin{tabular}{l c c}
        \hline
           Category                 & Percentage \\
        \hline              
        Image-to-Text               & 50\%    \\
        Annotation                  & 30\%    \\
        Dense Passage Retrieval     & 24\%    \\
        Generative QA               & 22\%   \\

        \hline
    \end{tabular}
    \caption{Ablation on Different $k$ in Beam-Search Decoding.}
    \label{tbl:error_analysis}
\end{table}

We can observe that the first contributing factor to the errors is in image-to-text transformation (50\%). The second category is the answer annotation ambiguity (30\%), where the predicted answers are reasonable according to human judgement, but do not match any ground-truth answers. There are also failures related to dense passage retrieval (24\%) and generative QA model (22\%). For more details of each error category, please see the examples in page 5-6.


\begin{figure*}[t!]
    \centering
    \includegraphics[width=1.0\linewidth]{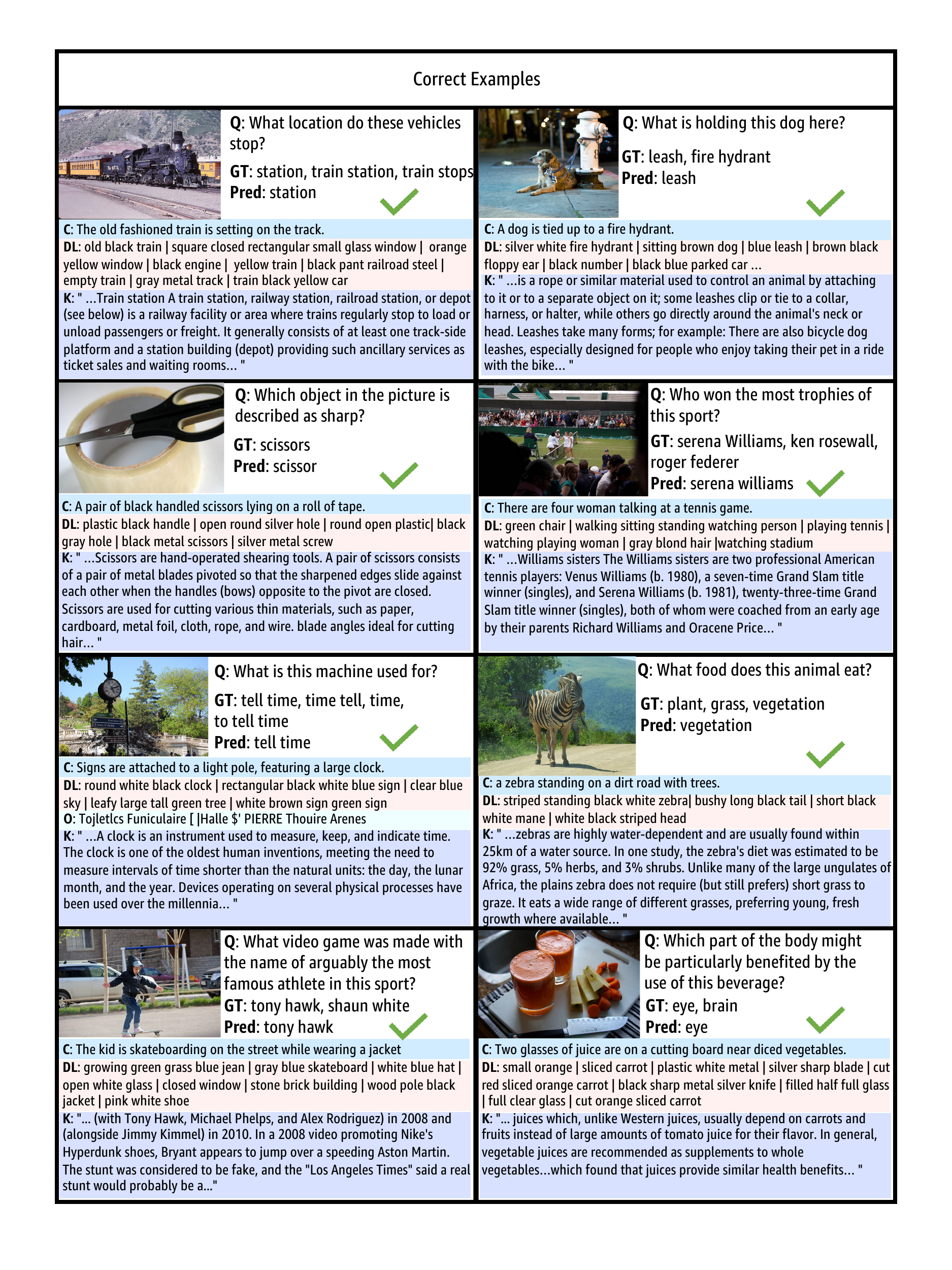}
    \label{fig:supp:correct_12}
\end{figure*}

\begin{figure*}[t!]
    \centering
    \includegraphics[width=1.0\linewidth]{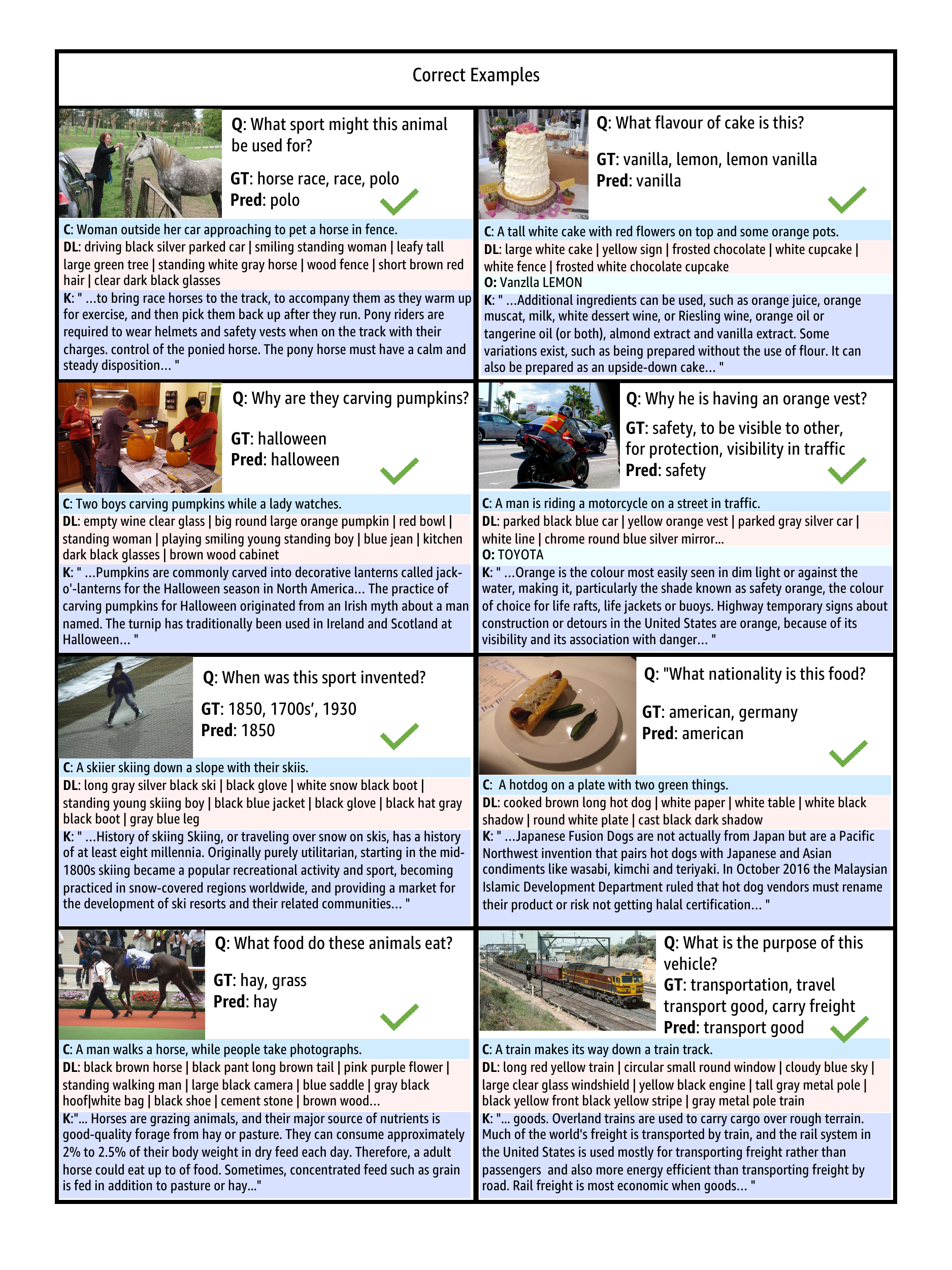}
    \label{fig:supp:correct_34}
\end{figure*}

\begin{figure*}[t!]
    \centering
    \includegraphics[width=1.0\linewidth]{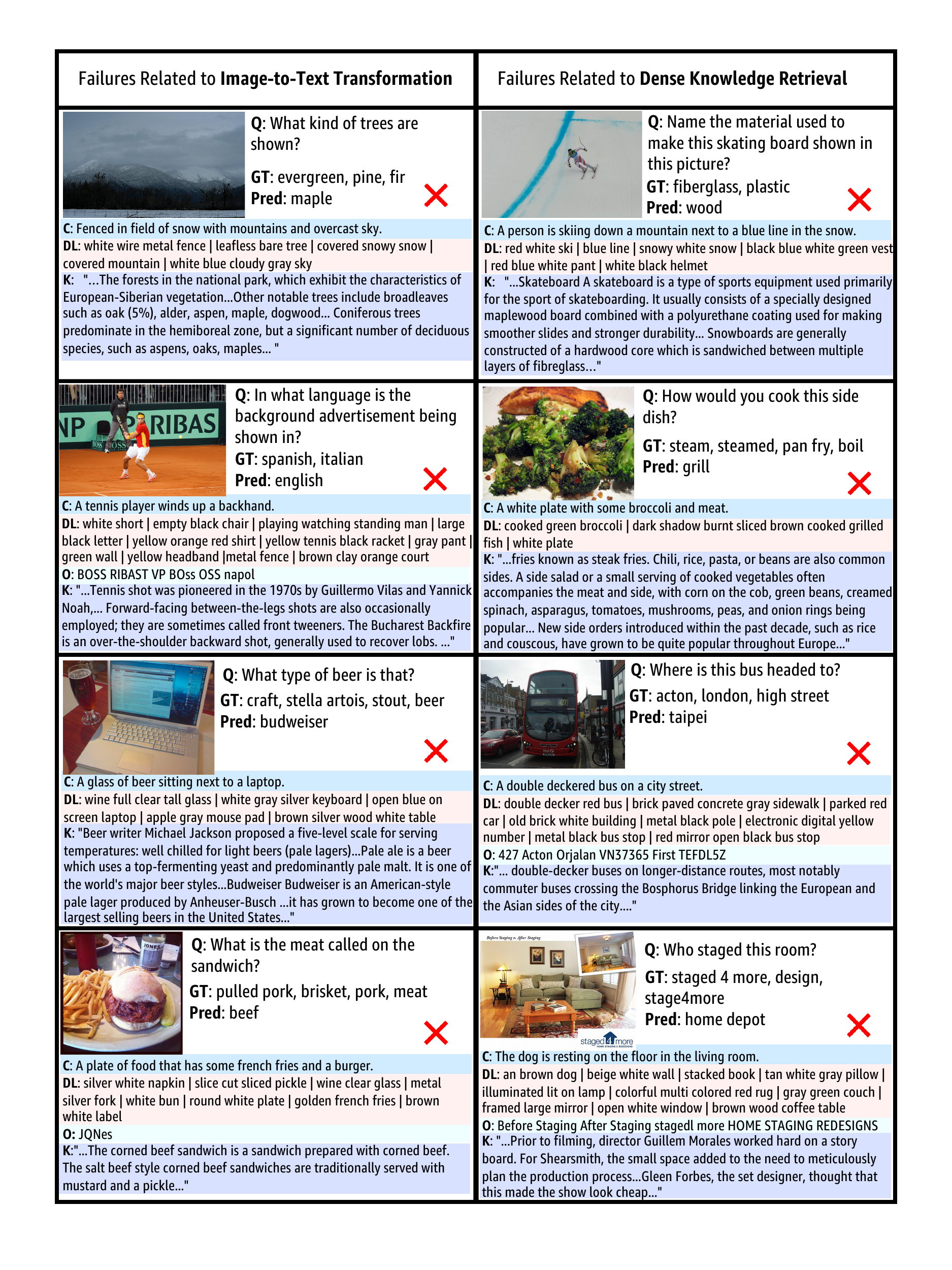}
    \label{fig:supp:error_12}
\end{figure*}

\begin{figure*}[t!]
    \centering
    \includegraphics[width=1.0\linewidth]{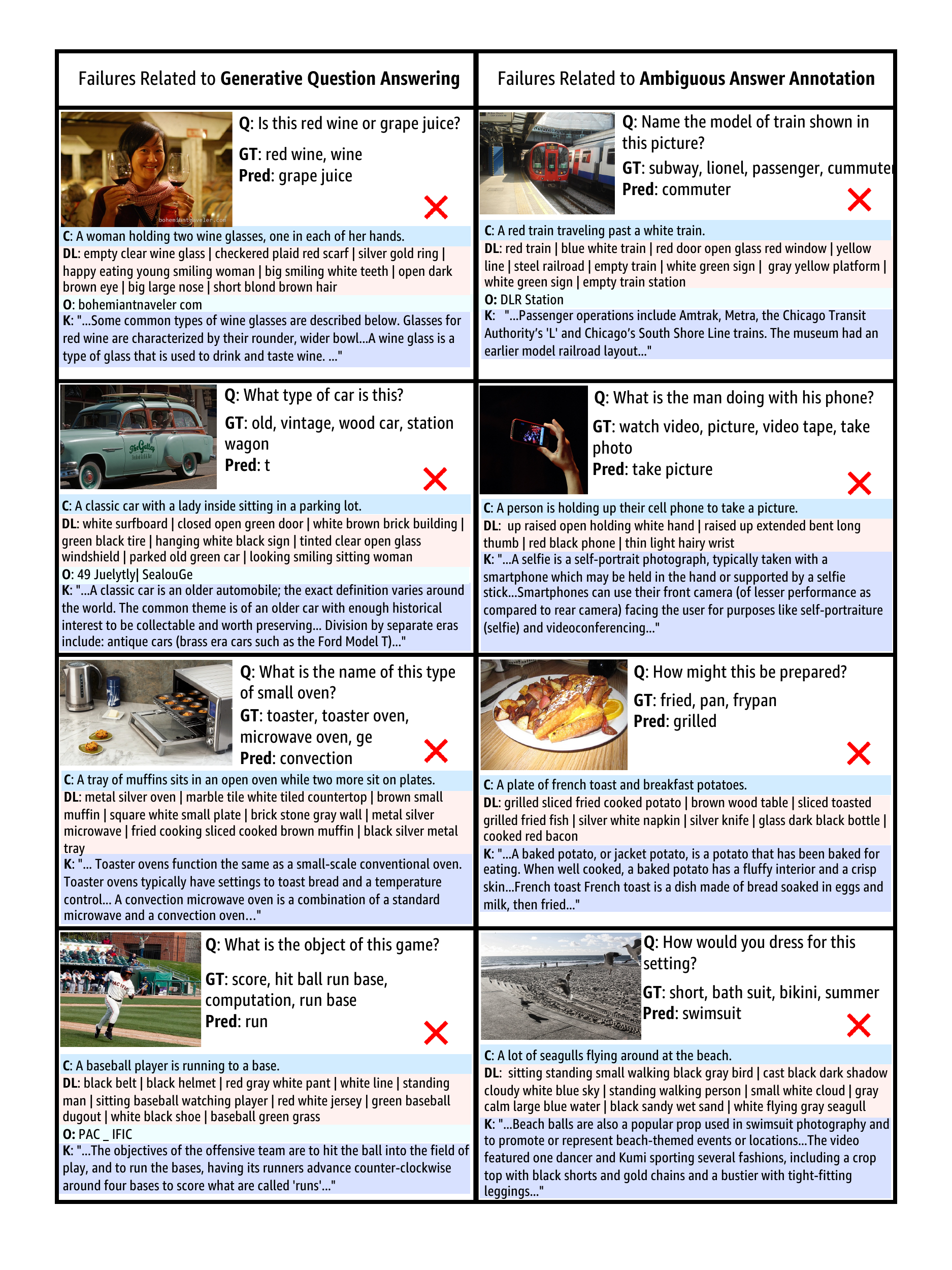}
    \label{fig:supp:error_34}
\end{figure*}

\end{document}